\documentclass[runningheads]{llncs}

\usepackage[T1]{fontenc}
\usepackage{graphicx,verbatim}

\usepackage{amsmath}
\usepackage{amssymb}
\usepackage{booktabs}
\usepackage{tabularx}

\usepackage{microtype} 

\usepackage{color}
\begin{document}
\title{LUMINA: Laplacian-Unifying Mechanism for Interpretable Neurodevelopmental Analysis via Quad-Stream GCN}
\titlerunning{LUMINA: Quad-Stream GCN for Neurodevelopmental Diagnosis}
%

\author{Minkyung Cha\inst{1} \and
Jooyoung Bae\inst{1} \and
Jaewon Jung\inst{1} \and
Ping Shu Ho\inst{2} \and
Ka Chun Cheung\inst{2} \and
Namjoon Kim\inst{1}}

\authorrunning{M. Cha et al.}

\institute{Seoul National University, Seoul, Republic of Korea \and
NVIDIA AI Technology Center HK, Hong Kong \\
\email{cmk4911@snu.ac.kr, domi37@snu.ac.kr, owenjaewon@snu.ac.kr, cliffh@nvidia.com, chcheung@nvidia.com, knj01@snu.ac.kr}}
  
\maketitle              
\begin{abstract}
Functional Magnetic Resonance Imaging(fMRI) has now become a  classic way for measuring brain activity, and recent trend is shifting toward utilizing fMRI brain data for AI-driven diagnosis. Given that the brain functions as not a discrete but interconnected whole, 
Graph Convolutional Network(GCN) has emerged as a dominant framework for such task, since they are capable of treating ROIs as 
dynamically interconnected nodes and extracting relational architecture between them. 
Ironically, however, it is the very nature of GCN's architecture that acts as an obstacle to its performance.
The mathematical foundation of GCN, effective for capturing global regularities, acts as a tradeoff; by smoothing features across the connected nodes repeatedly, traditional GCN tend to blur out the contrastive dynamics that might be crucial in identifying certain neurological disorders.

In order to break through this structural bottleneck, we propose LUMINA, a Laplacian-Unifying Mechanism for Interpretable Neurodevelopmental Analysis. Our model is a Quad-Stream GCN that employs a bipolar RELU activation and a dual-spectrum graph Laplacian filtering mechanism, thereby capturing heterogeneous dynamics that were often blurred out in conventional GCN. By doing so, we can preserve the diverse range and characteristics of neural connections in each fMRI data.
Through 5-fold cross validation on the ADHD200 ($N=144$) and ABIDE ($N=579$) dataset, LUMINA demonstrates stable diagnostic performance in two of the most critical neurodevelopmental disorder in childhood, ADHD and ASD, outperforming existing models with an accuracy of $84.66\%$ and $88.41\%$ each.

\keywords{Childhood Neurodevelopmental Disorder \and Graph Convolutional Networks \and Laplacian Filtering \and View Attention.}
\end{abstract}
\section{Introduction}
\subsection{fMRI as a Biomarker for Neurodevelopmental Disorder} 
ADHD (Attention Deficit Hyperactivity Disorder) and ASD (Autism Spectrum Disorder) are one
of the most common neurodevelopmental disorders in children\cite{ref1}.  
Timely therapeutic interventions could significantly improve a child's cognitive function and the social life afterwards, but not all children with ADHD or ASD are diagnosed at the right time, since current diagnosis relies heavily on clinical examination\cite{ref2,ref3}.
The difficulty of diagnosis escalates even further for individuals with relatively mild severity, since it's easy for them to unconsciously conceal their symptoms and blend in, a phenomenon called as social masking\cite{ref4}. Due to these diagnostic challenges, recent focus  has shifted toward the intrinsic brain activity itself measured by fMRI, rather than the clinical symptoms produced by it\cite{ref5}. fMRI measures the brain activity through the Blood-Oxygen-Level-Dependent(BOLD) signal based on the fact that on the fact that activated neurons induce increased local oxygen from the bloodstream\cite{ref6}. Among subtypes of fMRI, resting-state fMRI (rs-fMRI) is often utilized to capture the intrinsic brain dynamics in a natural condition\cite{ref7}. We focused on the rs-fmri of ADHD and ASD, and aimed to build a diagnostic model that could improve the early identification of childhood neurodevelopmental disorders.

\subsection{Graph Neural Networks}
For analysis, fMRI data are typically mapped onto a pre-defined brain atlas, where voxel signals are grouped and averaged within each ROIs. Pairwise correlations are then computed to establish functional connectivity between them\cite{ref8}. 
Graph Neural Networks (GNNs) have been widely adopted to process these functional connectomes, treating each ROI as a node and the pairwise correlations as edge weights. Most GNNs use the standard message passing mechanism based on the architecture of Graph Convolutional Network (GCN).\\
The initial connectome contains both positive and negative correlations, but standard message-passing mechanisms in GCN generally assume non-negative adjacency matrices. Therefore, existing approaches tend to apply absolute values ($|R|$) or positive thresholding ($\max(0, R)$) to the correlation matrix before constructing the graph\cite{ref:Chu}. 
As a result, the division between positive and negative correlationis erased; the bipolar nature of functional connectivity is discarded at this point. \\
One more limitation of existing GNN stems from the spectral properties of standard graph convolutions. For a graph with an adjacency matrix $A$ and a diagonal degree matrix $D$, the thoretic idea of convolutional kernel in standard GNN could be expressed as:
\begin{equation}
I + D^{-1/2} A D^{-1/2}
\end{equation} 
where $I$ is the identity matrix\cite{ref:math}. 
This acts inherently as a low-pass filter, since it performs a weighted averaging of features of a node and its connected neighbors\cite{ref:MedImage}. And when it's applied repeatedly, information in nodes with relatively high spatial frequency is oversmoothed and attenuated.
Standard GNNs are therefore structurally limited on their ability to detect heterogeneous characteristics of brain connectivity.

\section{Methodology}
\subsection{overview}
\begin{figure}
    \centering
    \includegraphics[width=1\linewidth]{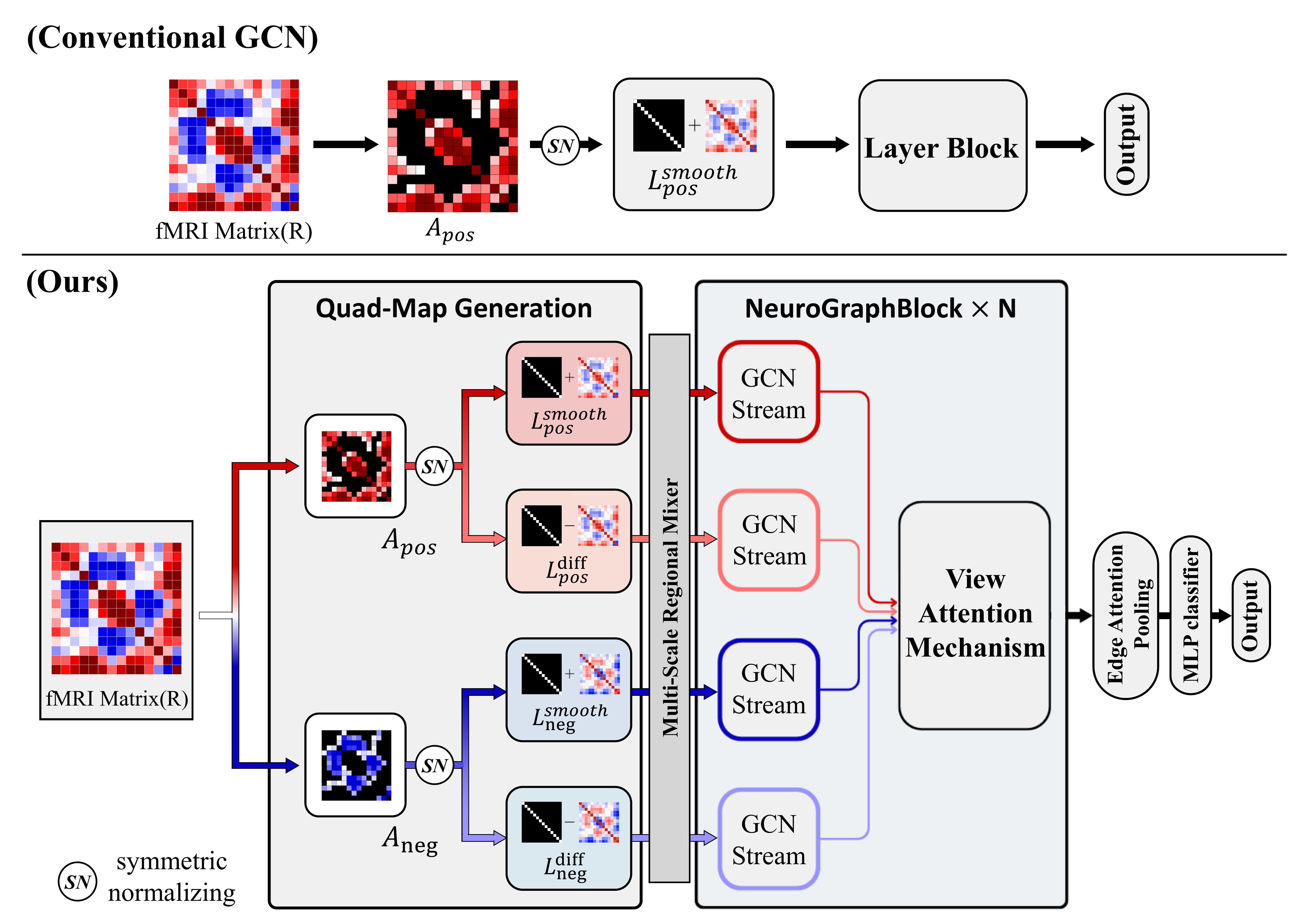}
    \caption{\textbf{Overview of the LUMINA framework.}
    Unlike conventional GCN(Top), our model(Bottom) incorporates a quad-stream input to capture diverse connectivity patterns. After processing by the Multi-Scale Regional Mixer and multiple GCN streams, they are passed through attention mechanism for final classification.}
    \label{fig:1}
\end{figure}

To address the current bottleneck, we propose LUMINA, an fMRI-based clinical diagnosis model designed to preserve the heterogeneous neural connections. The architecture of our model is outlined in Fig.~\ref{fig:1}; Conventional GCN typically utilizes only the absolute or positive values of adjacency matrices processed through a monolithic low-pass filter, but LUMINA segregates the input fMRI matrix ($R$) into positive and negative components ($A_{pos}$, $A_{neg}$), which are then transformed into four distinct streams ($L_{pos}^{smooth}$, $L_{pos}^{diff}$, $L_{neg}^{smooth}$, $L_{neg}^{diff}$). These maps are processed through the Multi-Scale Regional Mixer and multiple GCN streams, integrated by a View Attention Mechanism for final classification.

\subsection{Breakdown of LUMINA framework}
\subsubsection{Laplacian Filters and Quad-stream generation} 

Let ${X}^{(i)} \in \mathbb{R}^{N \times T}$ denote the fMRI time-series data for the $i$-th subject, where $N=111$ represents the number of Regions of Interest (ROIs) defined by the HO atlas, and $T=140$ is the number of time points. We first construct a functional connectivity matrix ${R}^{(i)} \in \mathbb{R}^{N \times N}$ by computing the Pearson correlation coefficient between all pairs of ROIs. For any two regions $u$ and $v$, the correlation $R_{uv}$ is defined as:

\begin{equation}
R_{uv} = \frac{\sum_{t=1}^{T} (x_{u,t} - \bar{x}_u)(x_{v,t} - \bar{x}_v)}{\sqrt{\sum_{t=1}^{T} (x_{u,t} - \bar{x}_u)^2} \sqrt{\sum_{t=1}^{T} (x_{v,t} - \bar{x}_v)^2}}
\end{equation}
\\
This matrix ${R}^{(i)}$ serves as the initial input feature for our model.\\
To preserve each distinct connection, we decompose ${R}$ into positive and negative components using the Rectified Linear Unit (ReLU) twice:
\begin{equation}
{A}_{pos} = \text{ReLU}({R}), \quad {A}_{neg} = \text{ReLU}(-{R})
\end{equation}
Next, we construct two types of spectral filters to capture the full spatial frequency. In Graph Signal Processing (GSP), the normalized graph Laplacian, defined as $L_{sym} = I - \hat{A}$ (where $\hat{A} = D^{-1/2} A D^{-1/2}$ is the symmetric normalized adjacency matrix and $D$ is the degree matrix), serves as an operator whose eigenvalues correspond to graph frequencies. In LUMINA, we first formulate a Smoothing filter as $L^{smooth} = I + \hat{A}$. This operation approximates a first-order low-pass filter on the graph spectral domain\cite{ref9}. 
Second, we formulate a Differential filter defined as $L^{diff} = I - \hat{A}$. This acts as a high-pass filter that amplifies the differences between connected nodes. Applying these two filters to the previously created components yields the following four equations:

\begin{equation}
\begin{aligned}
L_{pos}^{smooth} &= I + \hat{A}_{pos}, \quad &L_{pos}^{diff} &= I - \hat{A}_{pos} \\
L_{neg}^{smooth} &= I + \hat{A}_{neg}, \quad &L_{neg}^{diff} &= I - \hat{A}_{neg}
\end{aligned}
\end{equation}
\\
Consequently, this generates a set of quad-stream spatial priors, expressed as:
\begin{equation}
\mathcal{L} = \{ L_{pos}^{smooth}, L_{pos}^{diff}, L_{neg}^{smooth}, L_{neg}^{diff} \}
\end{equation}
where both positive/negative connection and high/low spatial frequency are conserved.
\subsubsection{NeuroGraphBlock and Final Classification} 
The next stage of LUMINA consists of an initial embedding layer, stacked NeuroGraph Blocks, and a final classification head. We first employ a Multi-Scale Regional Mixer, using 1D convolutions with dilation rates of ($d=\{1, 2, 5\}$) to extract the regional context between organized ROIs. Next, we process the node features ${H}$ through four parallel GCN streams, each guided by one of the maps in 
$\mathcal{L}$ from Eq. 5. \\
The node representation from the $k$-th prior ($k \in \{1, \dots, 4\}$) in $\mathcal{L}$ is denoted as:
\begin{equation}
Z_k = \mathcal{L}_k H W_k
\end{equation}
where $Z_k \in \mathbb{R}^{N \times d_{out}}$, $\mathcal{L}_k \in \mathbb{R}^{N \times N}$, $H \in \mathbb{R}^{N \times d_{in}}$, and $W_k \in \mathbb{R}^{d_{in} \times d_{out}}$.

Instead of a simple concatenation, we introduce a \textit{View Attention} mechanism to weight the importance of each view. The attention score $e_k$ for the $k$-th view is computed via a two-layer neural network:
\begin{equation}
e_k = \tanh(Z_k w_v + b_v) w_{attn}
\end{equation}
where $w_v$, $w_{attn}$, and $b_v$ are learnable parameters with $b_v$ denoting a bias term shared across all $N$ instances via broadcasting, and the respective dimensions are $e_k \in \mathbb{R}^{N \times 1}$, $w_v \in \mathbb{R}^{d_{out} \times d_{hidden}}$, $b_v \in \mathbb{R}^{1 \times d_{hidden}}$, and $w_{attn} \in \mathbb{R}^{d_{hidden} \times 1}$.

The final attention weights $\alpha_k$ are obtained by applying a softmax function across the four views:
\begin{equation}
\alpha_k = \frac{\exp(e_k)}{\sum_{j=1}^{4} \exp(e_j)}
\end{equation}
The  representation is then computed as the weighted sum of the four views:
\begin{equation}
{H}_{spatial} = \sum_{k=1}^{4} \alpha_k {Z}_k
\end{equation}
Finally, an Edge Attention Pooling layer aggregates the node-level features into a graph-level representation vector, which goes through MLP for final classification.
\section{Experiments and Results}

\subsection{Experimental Setup}
\subsubsection{Datasets.} For evaluation, we used rs-fMRI data from two public repositories: the Autism Brain Imaging Data Exchange (ABIDE) and the ADHD-200. We employed the preprocessed data of Harvard-Oxford (HO) atlas generated by the Configurable Pipeline for the Analysis of Connectomes (ABIDE), or the Athena pipeline (ADHD-200). Each of the 5-fold dataset was adjusted to maintain an balanced ratio in both labels(1:1) and data sources. The final cohort consisted of N=579 (ABIDE) and N=144 (ADHD-200) subjects.
\subsubsection{Baseline Models for Comparison.} For general graph-based baselines, we employed Graph Convolutional Networks (GCN) and Graph Attention Networks (GAT). The input functional connectomes were transformed using absolute values ($|R|$). For domain-specific baselines designed for connectome analysis, we employed BrainNetCNN and BrainGNN. 

\subsubsection{Implementation Details.} 
The hyperparameters for LUMINA were set as follows: $d_{out}=d_{in}=d_{hidden}=48$, number of layers $L=3$, learning rate $\eta=2e-4$ with AdamW optimizer, and weight decay $\lambda=1e-2$. 

\subsection{Performance Evaluation}
Table~\ref{tab:main} summarizes the performance comparison of LUMINA against 4 architectures. Our model achieved a mean classification accuracy of 84.66\% and 88.41\% in ADHD200 and ABIDE respectively, outperforming all baseline models.

\begin{table}[ht]
\centering
\caption{Classification performance of LUMINA and four baseline models. The best results are highlighted in bold.}
\label{tab:main}
\renewcommand{\arraystretch}{1} 
\begin{tabular*}{\textwidth}{@{\extracolsep{\fill}}llccc}
\toprule
\textbf{Dataset} & \textbf{Model} & \textbf{Accuracy (\%)} & \textbf{AUC} & \textbf{F1-Score} \\
\midrule
\textbf{ADHD-200} 
&GCN         & $81.71 \pm 3.50$ & $0.819 \pm .058$ & $0.796 \pm .016$ \\
&GAT         & $83.04 \pm 8.70$ & $0.839 \pm .147$ & $0.816 \pm .095$ \\
&BrainNetCNN & $78.44 \pm 7.57$ & $0.794 \pm .092$ & $0.762 \pm .075$ \\
&BrainGNN    & $74.23 \pm 10.56$ & $0.735 \pm .122$ & $0.594 \pm .344$ \\
&\textbf{LUMINA(ours)} & $\mathbf{84.66 \pm 9.53}$ & $\mathbf{0.861 \pm .097}$ & $\mathbf{0.849 \pm .103}$ \\
\midrule
\textbf{ABIDE} 
&GCN         & $78.19 \pm 3.80$ & $0.839 \pm .040$ & $0.775 \pm .046$ \\
&GAT         & $83.91 \pm 1.26$ & $0.888 \pm .032$ & $0.838 \pm .017$ \\
&BrainNetCNN & $82.00 \pm 2.26$ & $0.903 \pm .036$ & $0.810 \pm .029$ \\
&BrainGNN    & $62.30 \pm 13.04$ & $0.703 \pm .151$ & $0.371 \pm .331$ \\
&\textbf{LUMINA(ours)} & $\mathbf{88.41 \pm 1.57}$ & $\mathbf{0.936 \pm .018}$ & $\mathbf{0.885 \pm .016}$ \\
\bottomrule
\end{tabular*}
\end{table}
\subsection{Ablation Study}
We performed an ablation study to evaluate the individual contributions of LUMINA's each component(Table~\ref{tab:ablation_final_fixed}). We first tried replacing bidirectional RELU with absolute values $|R|$(\textit{w/o B/R}). The accuracy remained roughly the same in ABIDE, but decreased by 25.76\% in ADHD200. The removal of Dual-stream Laplacian(\textit{w/o D/L}) also led to similar result. The removal of NeuroGraphblock(\textit{w/o N/G}) led to decrease in performance in both datasets, by 21.96\%(ADHD200) and 18.01\% (ABIDE). The final removal of all 3 key components of LUMINA(Baseline) resulted in decrease by 21.26\%(ADHD200) and 18.81\%(ABIDE) respectively.

\newcolumntype{C}{>{\centering\arraybackslash}X}

\begin{table}[ht]
\centering
\caption{Ablation study of LUMINA}
\label{tab:ablation_final_fixed}
\setlength{\tabcolsep}{4pt}
\renewcommand{\arraystretch}{1.2}
\fontsize{8pt}{9.6pt}\selectfont

\begin{tabularx}{\textwidth}{l ccc X X X} 
\toprule
\textbf{Variant} & \multicolumn{3}{c}{\textbf{Components}} & \multicolumn{3}{c}{\textbf{Metrics}} \\
\cmidrule(lr){2-4} \cmidrule(lr){5-7}
& \textbf{B/R} & \textbf{D/L} & \textbf{N/G} & \textbf{Acc (\%)} & \textbf{AUC} & \textbf{F1} \\
\midrule
\textit{Dataset: ADHD-200} & & & & & & \\
\midrule
Baseline                & & & & $63.4 \pm 3.6$ & $0.625 \pm .054$ & $0.618 \pm .071$ \\
LUMINA w/o B/R          & & \checkmark & \checkmark & $58.9 \pm 4.8$ & $0.584 \pm .053$ & $0.585 \pm .045$ \\
LUMINA w/o D/L          & \checkmark & & \checkmark & $60.9 \pm 2.8$ & $0.608 \pm .040$ & $0.619 \pm .031$ \\
LUMINA w/o N/G          & \checkmark & \checkmark & & $62.7 \pm 4.9$ & $0.622 \pm .048$ & $0.602 \pm .064$ \\
\textbf{LUMINA (full)}  & \checkmark & \checkmark & \checkmark & $\mathbf{84.7 \pm 9.5}$ & $\mathbf{0.861 \pm .097}$ & $\mathbf{0.849 \pm .103}$ \\
\midrule
\textit{Dataset: ABIDE} & & & & & & \\
\midrule
Baseline                & & & & $69.6 \pm 0.8$ & $0.724 \pm .016$ & $0.698 \pm .029$ \\
LUMINA w/o B/R          & & \checkmark & \checkmark & $88.4 \pm 4.3$ & $0.934 \pm .027$ & $0.885 \pm .043$ \\
LUMINA w/o D/L          & \checkmark & & \checkmark & $88.2 \pm 3.9$ & $0.932 \pm .034$ & $0.884 \pm .040$ \\
LUMINA w/o N/G          & \checkmark & \checkmark & & $70.4 \pm 1.3$ & $0.728 \pm .022$ & $0.709 \pm .016$ \\
\textbf{LUMINA (full)}  & \checkmark & \checkmark & \checkmark & $\mathbf{88.4 \pm 1.6}$ & $\mathbf{0.936 \pm .018}$ & $\mathbf{0.885 \pm .016}$ \\
\bottomrule
\multicolumn{7}{l}{\scriptsize * B/R: Bidirectional ReLU, D/L: Dual-stream Laplacian, N/G: NeuroGraph Block.}
\end{tabularx}
\end{table}

\subsection{Post-hoc analysis}
\begin{figure}
    \centering
    \includegraphics[width=1\linewidth]{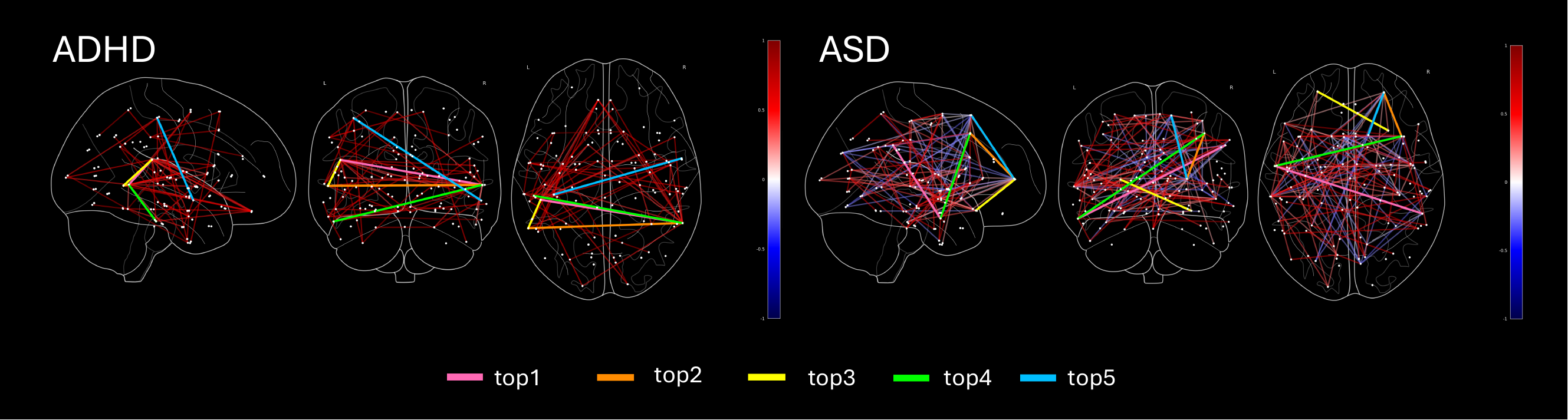} 
    \caption{\textbf{Post-hoc analysis of ADHD and ASD via the LUMINA framework.} Top 1\% ($\approx 61$) connections with highest IGs are shown as lines, with top 5 highlighted in distinct colors. The rest are visualized on a colorscale from blue ($-1$) to red ($+1$).}
    \label{fig:2}
\end{figure}

We performed post-hoc analysis via Integrated Gradients (IG) to identify the underlying factors contributing to the model's decision\cite{ref:IG}. 
We extracted the $111 \times 111$ attribution map from the trained LUMINA framework in the best fold of ADHD and ASD, and the results are shown in Fig.~\ref{fig:2} and Table~\ref{tab:top_connections}. In ADHD, all top 5 edges were hyperconnected, belonging in either Somatosensory(SMN) or Central Executive Network(CEN). In ASD, some hypoconnected edges were identified among the top 1\% but not in top 5, which were from CEN, Limbic Network(LN), and etc. \\
Post-hoc analysis reveals that the primary biomarkers diverge between the two disorders. The heavy reliance on hyperconnected edges within the SMN and CEN suggests that ADHD is strongly associated with the failure to suppress those networks, a conclusion consistent with established findings\cite{ref:ADHD}. Meanwhile in ASD, the mixed presence of connections hint that a more heterogeneous disruption of socio-emotional and cognitive integration may underlie.

\newcolumntype{L}{>{\raggedright\arraybackslash}X}

\begin{table}[ht]
\centering
\caption{Top 5 Functional Connections for ADHD and ASD Groups}
\label{tab:top_connections}
\setlength{\tabcolsep}{4pt}
\renewcommand{\arraystretch}{1.1}
\fontsize{8pt}{9.6pt}\selectfont

\begin{tabularx}{\textwidth}{@{} c cc L L @{}}
\toprule
\textbf{Rank} & \textbf{R} & \textbf{IG} & \textbf{Node A (Network)} & \textbf{Node B (Network)} \\
\midrule
\multicolumn{5}{l}{\textit{Top 5 ADHD Connections}} \\
\midrule
1 & 0.67 & 0.0081 & Right Parietal Operculum Cortex (SomatoMotor B) & Left Middle Temporal Gyrus; temporooccipital part (Control C) \\
2 & 0.63 & 0.0056 & Right Middle Temporal Gyrus; temporooccipital part (Control C) & Left Middle Temporal Gyrus; temporooccipital part (Control C) \\
3 & 0.53 & 0.0051 & Right Parietal Operculum Cortex (SomatoMotor B) & Right Middle Temporal Gyrus; temporooccipital part (Control C) \\
4 & 0.61 & 0.0051 & Right Inferior Temporal Gyrus; posterior division (Control C) & Left Middle Temporal Gyrus; temporooccipital part (Control C) \\
5 & 0.75 & 0.0048 & Right Postcentral Gyrus (SomatoMotor A) & Left Superior Temporal Gyrus; anterior division (TempPar) \\

\midrule
\multicolumn{5}{l}{\textit{Top 5 ASD Connections}} \\
\midrule
1 & 0.58 & 0.0246 & Left Supramarginal Gyrus; posterior division (Ventral Attention) & Right Middle Temporal Gyrus; anterior division (TemporoParietal) \\
2 & 0.36 & 0.0157 & Left Frontal Pole (Limbic B) & Left Middle Frontal Gyrus (Control B) \\
3 & 0.66 & 0.0150 & Left Frontal Orbital Cortex (Limbic B) & Right Frontal Pole (Limbic B) \\
4 & 0.37 & 0.0140 & Left Middle Frontal Gyrus (Control B) & Right Middle Temporal Gyrus; anterior division (TempPar) \\
5 & 0.50 & 0.0140 & Left Frontal Pole (Limbic B) & Left Superior Frontal Gyrus (Control A) \\
\bottomrule
\multicolumn{5}{l}{
  \shortstack[l]{ 
    * R: Correlation value, IG: Integrated Gains.}
}
\end{tabularx}
\end{table}

\section{Conclusion}
We propose LUMINA, a quad-stream GCN-based framework for fMRI based clinical diagnosis that preserves the heterogeneous neural connection through unique processing with bidirectional RELU and dual laplacian filtering. Comparison against baseline models demonstrates the stable performance of LUMINA across subtypes of neurodevelopmental disorders, and post-hoc analysis on the model's performance identified the key networks and connections that aligned with established findings, suggesting the potential as an effective, interpretable diagnostic model. We expect the proposed framework to act as an insight into relatively unexplored neurological disorders in future research.

%
%

\end{document}